\pdfoutput=1
\documentclass[10pt,twocolumn,letterpaper]{article}
\usepackage{cvpr}
\usepackage{times}
\usepackage{epsfig}
\usepackage{graphicx}
\usepackage{amsmath}
\usepackage{amssymb}
\usepackage{subfigure} 
\usepackage{pfnote}

\newcommand{\Eref}[1]{Equation~{\ref{#1}}}
\newcommand{\Tref}[1]{Table~{\ref{#1}}}
\newcommand{\Fref}[1]{Figure~{\ref{#1}}}
\newcommand{\Sref}[1]{Section~{\ref{#1}}}

\usepackage[breaklinks=true,bookmarks=false]{hyperref}

\cvprfinalcopy 

\def\cvprPaperID{1784} 

\begin{document}

\title{Learning to Deblur and Generate High Frame Rate Video\\ with an Event Camera}
\author{Haoyu Chen, Minggui Teng, Boxin Shi, Yizhou Wang, Tiejun Huang\\
Peking University, Beijing, China\\
{\tt\small \{haoyu\_chen, minggui\_teng, shiboxin, yizhou.wang, tjhuang\}@pku.edu.cn}
}
\maketitle

\begin{abstract}
   Event cameras are bio-inspired cameras which can measure the change of intensity asynchronously with high temporal resolution. One of the event cameras' advantages is that they do not suffer from motion blur when recording high-speed scenes. In this paper, we formulate the deblurring task on traditional cameras directed by events to be a residual learning one, and we propose corresponding network architectures for effective learning of deblurring and high frame rate video generation tasks. We first train a modified U-Net network to restore a sharp image from a blurry image using corresponding events. Then we train another similar network with different downsampling blocks to generate high frame rate video using the restored sharp image and events. Experiment results show that our method can restore sharper images and videos than state-of-the-art methods.
\end{abstract}

\begin{figure*}
\centering
\includegraphics[width=1\textwidth]{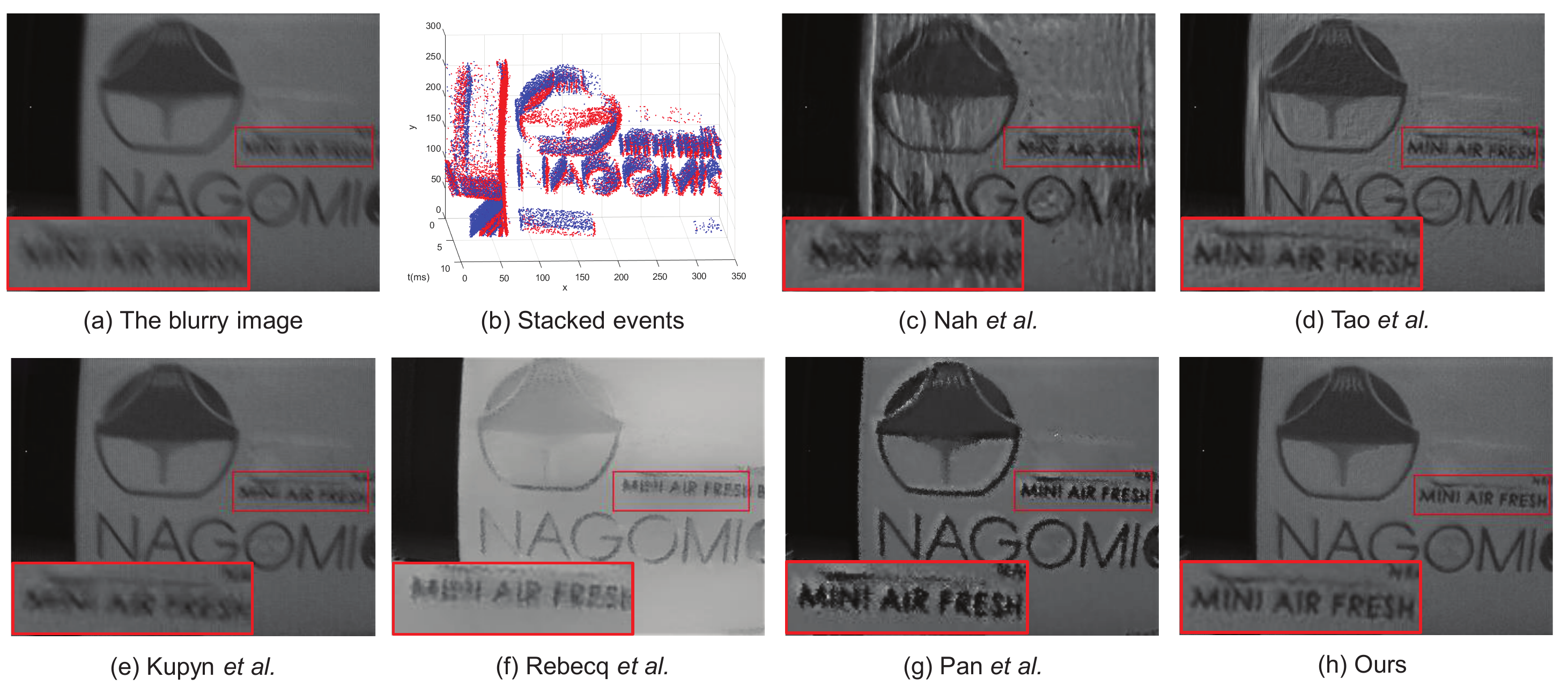}
\centering
   \caption{An example of deblurring result. (a) The input blurry image. (b) The corresponding event data. (c)$\sim$(h) Deblurring results of Nah \etal~\cite{Nah2017Deep}, Tao \etal~\cite{Tao2018Scale}, Kupyn \etal~\cite{Kupyn_2019_ICCV}, Rebecq \etal~\cite{Rebecq2019Highspeed} (using only events), Pan \etal~\cite{Pan2019EDI} and our method.  
   }
\label{fig:real data example1}
\end{figure*}
\section{Introduction}
Event cameras, such as the Dynamic Vision System (DVS)~\cite{Lichtsteiner2008A} and the Dynamic and Active-pixel Vision Sensor (DAVIS)~\cite{Brandli2014A}, are bio-inspired sensors that asynchronously detect the change of log intensity at each pixel independently, comparing to traditional cameras that sample intensity at each pixel during the exposure time and form a picture. If the intensity change reaches a threshold, the camera will trigger an event, $\textbf{e}=\{\textbf{u},t,p\}$, in which $\textbf{u}=(x,y)^T$ is the coordinate of the pixel, $t$ is the timestamp when the event is generated, and $p$ is the polarity of the event. For a pixel $\textbf{u}$ whose intensity is $\textbf{I}(t_0)$ in an  event camera, it measures the change of intensity. If at time $t$ the log intensity ($\textbf{L}=\log\textbf{I}$) change $|\textbf{L}(\textbf{u},t)-\textbf{L}(\textbf{u},t_0)|\geq \delta$, it will output an event $\textbf{e}=(\textbf{u},t,p), p = 1$ (or $-1$) if the difference is greater (or less) than 0. So outputs of event cameras are not frames but a sequence of events with coordinate, timestamp, and polarity information.  

Compared with traditional cameras, event cameras have several advantages such as higher temporal resolution (around 1$\mu$s) and higher dynamic range (140dB). As a result, event cameras would not suffer from many problems that traditional cameras have. An important one of them is the motion blurring of images whose removal is still a challenging problem in computer vision research.

Motion blur is the result of the relative motion between the camera and the scene during the integration time of image capturing, such as camera shake or object motion. Traditional methods to restore a sharp image from a blurry one apply various constraints to model characteristics of blur ($e.g.$, non-blind uniform/blind uniform/non-uniform), and utilize different natural image priors to regularize the solution space~\cite{Chan1998Total, Cho2009Fast, Weisheng2013Nonlocally,Zhang2013Multi, Li2013Unnatural,Pan2014Deblurring}. Most of these methods involve intensive, sometimes heuristic, parameter-tuning and expensive computation. Recently, learning-based methods have also been proposed for deblurring. Early methods follow the idea of traditional methods and substitute some operators with learned modules~\cite{Jian2015Learning,Schuler2016Learning,Lei2016Learning}. Recent methods focus on designing end-to-end network structures for image deblurring~\cite{Nah2017Deep,ZhangPRSBL018,Tao2018Scale} and performing blind deblurring without estimating the blur kernels. 

Event cameras have been applied to various computer vision tasks by making compatible representations with image-based computer vision algorithms~\cite{Reinbacher2016Real,Barua2016Direct,Sironi2018HATS,Zhu2018Unsupervised}. It is natural to apply event
cameras for motion deblurring due to their high sensitivity to object motion. By connecting images with motion blur and events~\cite{Brandli2014Real,Scheerlinck2018Continuous,Pan2019EDI},  
the complementary information from two types of cameras could accomplish the motion deblurring task with higher performance and lower cost. 

In this paper, we propose a learning-based method combined with events to restore a sharp image from a blurry image and generate a high frame rate (HFR) video. Unlike existing works that generate sharp images using events directly, we formulate the deblurring task to be a residual learning one and propose a modified U-Net~\cite{Ronneberger2015Invited} architecture with DenseNet~\cite{Li2017Classification} blocks in which event stacks can be transformed to be a mask. By adding the mask to a blurry image, the sharp one will be restored. After the deblurring procedure, we can use the restored sharp image with events to generate HFR video using a similar residual learning network architecture, in which DenseNet blocks are replaced by Conv-LSTM blocks. The contributions of this paper can be summarized as:

\begin{itemize}
    \item We revisit the relationship between a blurry image and events, and propose a residual model suitable for learning image deblurring and HFR video generation.
    \item We propose a modified U-Net with a global residual connection which is particularly effective for our residual model and obtains a better result on deblurring than state-of-the-art methods (\Fref{fig:real data example1}). 
    \item We propose another residual network to recurrently generate the next clear frame guided by events and design a pipeline to concatenate several blurry images to an HFR video.
\end{itemize}
\section{Related Work}
In this section, we will review the previous image deblurring and event-based methods related to our work.
\subsection{Learning-based image and video deblurring}
As deep learning has shown its striking effects in solving computer vision problems such as object detection and segmentation, image and video deblurring also benefits from deep learning methods. Sun \etal~\cite{Jian2015Learning} proposed a deep learning approach to predicting the probabilistic distribution of motion blur at the patch level using a convolutional neural network and removed the motion blur by a non-uniform deblurring model using a patch-level image prior. 

Unlike the early method~\cite{Jian2015Learning} that replaced the calculation of blur kernel with a classification network, end-to-end methods generating deblurring images directly were proposed. Nah \etal~\cite{Nah2017Deep} proposed a multi-scale CNN that directly restores latent images without assuming any restricted blur kernel model without estimating explicit blur kernels. Further, Tao \etal~\cite{Tao2018Scale} proposed a multi-scale encoder-decoder network. Inspired by the coarse-to-fine scheme the blurry image was restored on different resolutions in the scale-recurrent structured network, which had a simpler network structure, a smaller number of parameters and was easier to train. Kupyn \etal~\cite{Kupyn_2019_ICCV} used another popular network structure, generative adversarial network (GAN), to perform the deblurring task. They formulated the deblurring task to be a residual learning one. Their method could restore the image using fewer calculation resources in a shorter time. Jin \etal~\cite{Jin2018Learning} proposed a network to extract video from a blurry image by restoring the middle sharp image and calculating temporal ambiguities. Zhang \etal~\cite{zhang2018adversarial} proposed a 3D convolution to both spatial and temporal domains to extract the motion features in the video and used a GAN to make the restored video sharper and more realistic.

\subsection{Event-based computer vision methods}
 Events have shown unique advantages in estimating optical flows. Bardow \etal~\cite{Bardow2016Simultaneous} aimed to simultaneously recover optical flow and HDR images employing minimization of a cost function that contains the asynchronous event data as well as spatial and temporal regularisation. Zhu \etal~\cite{Zhu2018Unsupervised} proposed a FlowNet-liked~\cite{Dosovitskiy2015FlowNet} network for unsupervised learning the optical flow, egomotion and depths only from the event stream. 

There are also several works trying to relate events with intensity frames and reconstruct high-quality images even high frame-rated videos. Reinbacher \etal~\cite{Reinbacher2016Real}  proposed a variational model enabling reconstruction of intensity images with an arbitrary frame rate in real-time via an event manifold induced by the relative timestamps of events.  Scheerlinck \etal~\cite{Scheerlinck2018Continuous} proposed a continuous-time formulation of event-based intensity estimation using complementary filtering to combine image frames with
events and obtained continuous-time image intensities. Mohammad \etal~\cite{Mohammad2019EventHDR} used event-based cGAN~\cite{DBLP:journals/corr/MirzaO14} to create images/videos from an adjustable portion of the event data stream based on the spatio-temporal intensity changes. 

Events are also used in the deblurring and high-speed video generation tasks because of the advantage of high temporal resolution and reliable motion information encoded in captured events. Pan \etal~\cite{Pan2019EDI} proposed an optimization method for estimating a single scalar variable, named the Event-based Double Integral (EDI) model, to restore a sharp image. Then an HFR video can be generated from the restored image using the optimized scalar, which is physically corresponding to the threshold of triggering the event. Rebecq \etal~\cite{Rebecq2019Highspeed} proposed a U-Net-like network with Conv-LSTM blocks and generated a video directly from encoded events in a spatio-temporal voxel grid. 

The input and output of our method is the same as~\cite{Pan2019EDI}, and our method is also inspired by the physical model of an event camera. However, our model suffers less from noisy events and avoids rapid accumulation of errors due to the stronger representation power of residual mode, as we will prove using extensive experiments.

\section{Proposed Method}
In this section, we will introduce our proposed learning-based methods for the deblurring task (\Sref{sec:formulation}) and the design methodology of the network (\Sref{sec:model 1}). Based on the output of the deblurring task, we propose a method to generate a sequence of frames using the restored image and related events (\Sref{sec:model 2}). We will discuss and compare our method with existing methods that restore and generate images using events (\Sref{sec:comparison}). We will also explain the architecture and training details of our networks respectively.

\subsection{Event-based residual image formation model}\label{sec:formulation}
\subsubsection{Image deblurring model}
To restore a blurry image using events from a DVS camera, we should build the connection between them.

Given the intensity of sharp images, as described in~\cite{Nah2017Deep}, blur accumulation process can be modeled as,
\begin{equation}\label{eq:blur}
\textbf{B} = \frac{1}{T-t_0}\int_{t_0}^{T}\textbf{I}(t)dt \approx \frac{1}{T-t_0}\sum_{t_0}^T\textbf{I}(t).    
\end{equation}

Supposing that $\epsilon(t_1,t)=\{\textbf{e}_i\}_{i=1}^n$ is the incoming event stream at a pixel from $t_1$ to $t$, then the latent image at $t$ can be expressed as 
\begin{equation}\label{eq:latent image}
    \textbf{I}(t) = \textbf{I}(t_1)\cdot \exp\left(\sum_{i=1}^n C_i\right),
\end{equation}
in which $C_i$ is the threshold of intensity change to trigger the $i$-th event $\textbf{e}_i$. Such a threshold has different values for positive and negative intensity change, and it is not a constant which generally follows a normal distribution over time~\cite{Rebecq18corl}. It is non-trivial to give an analytical expression for the change caused by $\textbf{e}_i$ at time $t$, so we donate it as $f(\textbf{e}_i,\textbf{I}(t))$. We expect to accurately estimate the change of an event using the local intensity and gradient information.

By combining \Eref{eq:blur} and \Eref{eq:latent image} we obtain
\begin{equation}\label{eq:our model}
    \textbf{B}=\frac{1}{T-t_0}\sum_{t_0}^T\textbf{I}(t_0)\cdot \prod_i f(\textbf{e}_i,\textbf{I}(t)).
\end{equation}

Taking the logarithm on the both sides of \Eref{eq:our model}, in which $\textbf{L} = \log \textbf{I}$, and rearranging the equation, yields
\begin{equation}\label{eq:To approx}
    \textbf{L}(t_0)=\log\textbf{B}-\log\left(\frac{1}{T-t_0}\sum_{t_0}^T\prod_i f(\textbf{e}_i,\textbf{I}(t))\right).
\end{equation}

While in the deblurring task, the intensity value of the sharp image $\textbf{I}(t)$ is unavailable. Actually, in the blurry image, gradient information can still be observed along the edge of the blurry part. So we approximately use $\textbf{B}$ as a proxy of $\textbf{I}(t)$ for $f(\textbf{e}_i,\textbf{I}(t))$ in \Eref{eq:To approx} as the following equation:
\begin{equation}\label{eq:result}
    \textbf{L}(t_0)=\log\textbf{B}-\log\left(\frac{1}{T-t_0}\sum_{t_0}^T\prod_i f(\textbf{e}_i,\textbf{B})\right).
\end{equation}

\subsubsection{HFR video generation model}
\Eref{eq:result} provides the mapping from a blurry image to a sharp image using a residual term encoded with events. Baesd on the restored shape image and subsequent events, we can generate HFR videos thanks to the high temporal resolution of events. Using the log representation of \Eref{eq:latent image}, we obtain the residual relationship between two adjacent frames:
\begin{equation}\label{eq:video}
    \textbf{L}(t_{k+1})=\textbf{L}(t_k)+\sum_{i=1}^{n}\log\left(f(\textbf{e}_i,\textbf{I}(t_k))\right),(k=0,\cdots,Q-1),
\end{equation}
where $\textbf{L}(t_0)$ is the restored image, and the Q generated images $\{\textbf{L}(t_1),\cdots,\textbf{L}(t_{Q})\}$ are combined to be an HFR video.

In the following subsections, we propose two corresponding network designs to learn deblurring using \Eref{eq:result} and HFR video generation using \Eref{eq:video}.

\begin{figure}
\begin{center}
\includegraphics[width=3.3in]{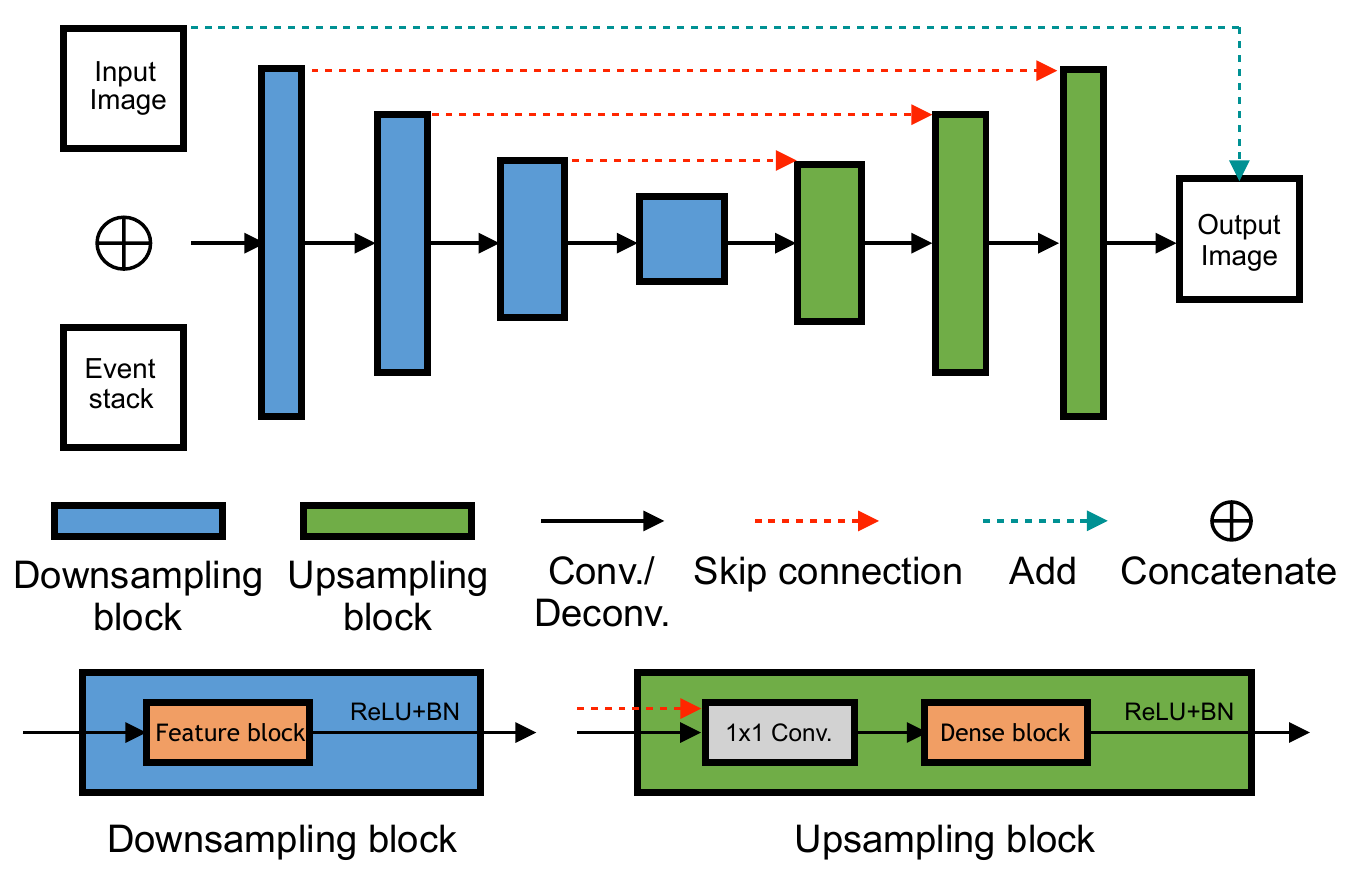} 
\end{center}
   \caption{The architecture of our event-based residual deblurring/HFR video generation net. The DenseNet/Conv-LSTM blocks are used in the downsampling procedure as the feature blocks. The input of the network is concatenation of binned events and an image, which is the blurry image/previous frame. The output is the sharp image/next frame.}
   \label{fig:U-Net_architecture}
\end{figure}

\subsection{Event-based residual deblurring net}\label{sec:model 1}
\paragraph{Network architecture.}
According to \Eref{eq:result}, our goal is restoring a sharp image using events. While the function $f(\textbf{e}_i,\textbf{B})$ is highly undetermined and hard to be explicitly described. So we design a learning-based method to learn such a function from training data. We propose a generation network whose architecture is shown in \Fref{fig:U-Net_architecture}. In addition, we introduce a global skip connection, which matches our formulation and makes the generation results better, then our network can be treated as a residual learning one.

As described above, given the blurry image and events, we denote our prediction as
\begin{equation}
F(\epsilon,\textbf{B},\theta)=\log\left(\frac{1}{T-t_0}\sum_{t_0}^T\prod_i f(\textbf{e}_i,\textbf{B})\right),    
\end{equation}
where $F(\epsilon,\textbf{B},\theta)$ is the output of our network trained with the event sequence $\epsilon$, blurry image $\textbf{B}$ and $\theta$, which consists of all network parameters to be learned. According to the results in \cite{Pan2019EDI}, features like gradient can guide the deblurring tasks effectively. So our network should learn the relationship between the event sequence and the change of intensity depending on the local information and features in the blurry image.  

Our network consists of two parts: the encoder part, which is used to extract features in the blurry images and stacked events frames, and the decoder part, which is used to fuse encoded blurry images with the encoded events.

The encoder part is designed based on DenseNet to extract local features, which may simply be the gradient of the blurry image and the gradient of the current estimate, possibly thresholded in flat regions~\cite{Li2010Two}. Recent works also show that DenseNet can be trained to extract high-level features in an image even better than the popular ResNet~\cite{huang2017densely}. We also train our encoder with stacked event frames to extract features from events in a neighborhood. To use the features extracted from our DenseNet based encoder, we replace the fully-connected layers in DenseNet model by a $3\times 3$ convolutional layer so we can input them into our decoder. 

After extracting features with our encoder, we design a feature fusion block in decoder of our network to fuse features extracted by the encoder from the image stream and event stream.
To fuse features and learn the change of intensity caused by events at a pixel, we use a $1\times 1$ convolutional layer to skip connect features in downsampling procedure with those in the upsampling procedure. Moreover, to fuse features depending on the local area of a pixel, we use a $3\times 3$ convolutional layer in the upsampling procedure. Our network can also filter the noise from input event sequence to a certain degree. Noisy event, which is sparse locally, will not cause effective activation because of the ReLU structure in the network.

\paragraph{Data input.}\label{sec:input}
The output of event cameras is a sequence consisting of event tuples $(\textbf{u},t,p)$. We need to find a representation of events so that we can feed them into our network. We adopt the representation in~\cite{Mohammad2019EventHDR} which constructs 3D event volume via merging and stacking the events within a small time interval. We apply this representation to our input of event sequences. Then events are concatenated with the blurry image as several channels of the input. Moreover, as described in~\cite{Mohammad2019EventHDR}, stacking events into multiple frames is better than one. More details of the data input will be shown in \Sref{sec:training details}.

\paragraph{Loss function.}\label{sec:loss}
We define our loss function as follows:
\begin{equation}\label{eq:loss function}
    \mathcal{L}(I_B,I_S) = \mathcal{L}_1(I_B,I_S) + \lambda\cdot \mathcal{L}_{PL}(I_B,I_S),
\end{equation}
where $I_S$ means generated images and $I_B$ means blurry images. $\mathcal{L}_1(I_B,I_S)$ is the $\mathcal{L}_1$-loss or MAE loss between $I_B$ and $I_S$. The perceptual loss $\mathcal{L}_{PL}$~\cite{Johnson2016Perceptual} is an improved kind of $\mathcal{L}_2$-loss based on VGG19 trained on ImageNet. It does not calculate the difference between generated images and target images but the sum of differences in the feature maps. Perceptual loss measures high-level perceptual and semantic differences between images to provide more robust constraints than $\mathcal{L}_2$-loss. The perceptual loss can be defined as  the following:
\begin{equation}
    \begin{split}
    &\mathcal{L}_{PL}(I_B,I_S) =\\ &\frac{1}{W_{i,j}H_{i,j}}\sum_{x=1}^{W_{i,j}}\sum_{y=1}^{H_{i,j}}(\phi_{i,j}(I_B)_{x,y}-\phi_{i,j}(I_S)_{x,y})^2,
    \end{split}
\end{equation}
where $\phi_{i,j}$ is the feature map obtained by the $j$-th convolution (after activation) before the $i$-th maxpooling layer, $W_{i,j}$ and $H_{i,j}$ are the dimensions of the feature maps.

\subsection{Event-based residual HFR video generation net}\label{sec:model 2}
\paragraph{Network architecture.}
Similar to deblur task, we can design a residual learning network to solve the problem in \Eref{eq:video} for HFR video generation. Given events and the previous frame, we denote the prediction as
\begin{equation}
    G(\epsilon(t_k,t_{k+1}),\textbf{L}(t_k),\theta)=\sum_{i=1}^{n}\log\left(f(\textbf{e}_i,\textbf{I}(t_k))\right).
\end{equation}
A straightforward design of network is using events and one frame as input and the output is the next frame. However, we experimentally find this design is hard to train and lead to poor performance because of rapid accumulation of error (the generated image quality drops catastrophically after the third frame). Considering this consequence, we propose our network to be a residual net, which also follows the video generation model we propose in \Sref{sec:formulation}. 

We still use the structure we proposed for the deblurring task as shown in \Fref{fig:U-Net_architecture}. While this time, we replace the DenseNet blocks in the downsampling procedures with Conv-LSTM~\cite{xingjian2015convolutional} blocks to learn the temporal information in the event sequences because temporal consistency is the key in video generation task, which is proved to be valid in~\cite{Rebecq2019Highspeed}. 

\paragraph{Data input.}
The inputs of our HFR video generation network are a sharp picture and events in the next period of time. The first sharp picture is the restored one using our event-based deblurring residual net. Then we will recurrently use the generated frame to be the input in the next step. As described in \Sref{sec:model 1}, we divide the input events into 6 bins. The loss function is also the same as \Eref{eq:loss function}. 

\begin{figure}[t]
\begin{center}
\includegraphics[width=3.3in]{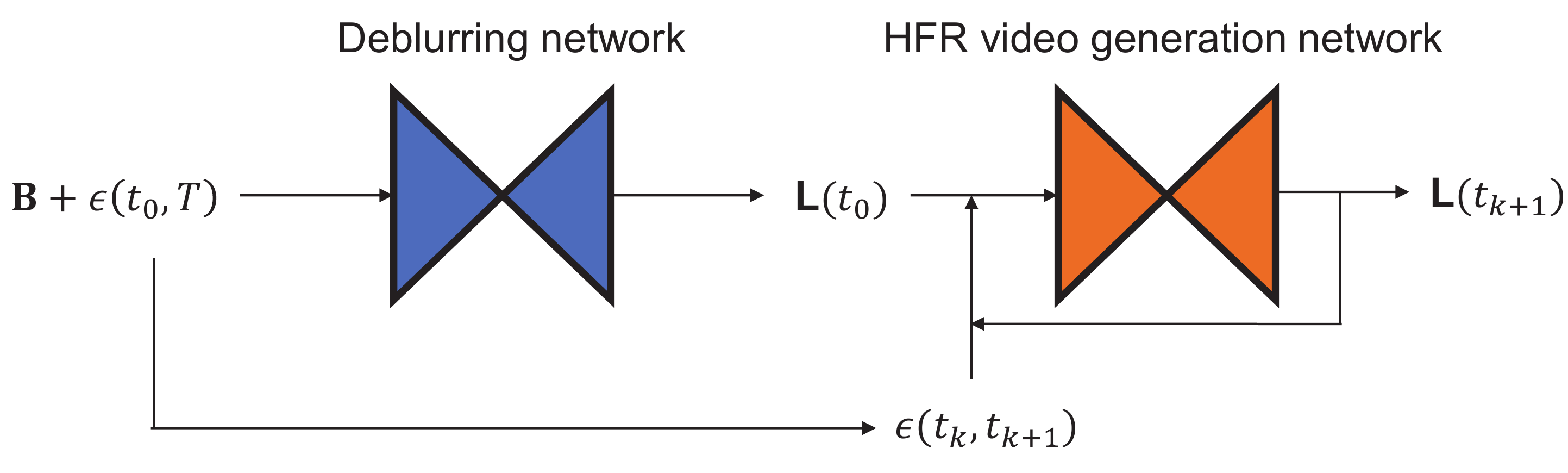} 
\end{center}
   \caption{The pipeline of our method to restore HFR video from a blurry image. At the first stage, we use our event-based deblurring residual net (\Fref{fig:U-Net_architecture} with DenseNet in downsampling blocks) with all the events $\epsilon(t_0,T)$ and the blurry image $\textbf{B}$ to obtain the sharp image $\textbf{L}(t_0)$. Next, we recurrently use our event-based HFR video generation residual net (\Fref{fig:U-Net_architecture} with Conv-LSTM in downsampling blocks) with part of the events $\epsilon(t_k,t_{k+1})$ and the output frame $\textbf{L}(t_k)$ to predict the next frame $\textbf{L}(t_{k+1})$.  
   }
\label{fig:Pipeline}
\end{figure}

The pipeline of our method to restore HFR video from a blurry image is described in \Fref{fig:Pipeline}. It consists of two stages. At the first stage, using events and blurry image, our event-based deblurring residual net restores a sharp image, which will be used as the first frame at the second stage. Then our event-based HFR video generation residual net recurrently uses part of the events and a sharp frame to generate the corresponding next frame. 

\begin{figure}[t]
\centering
\includegraphics[width=0.45\textwidth]{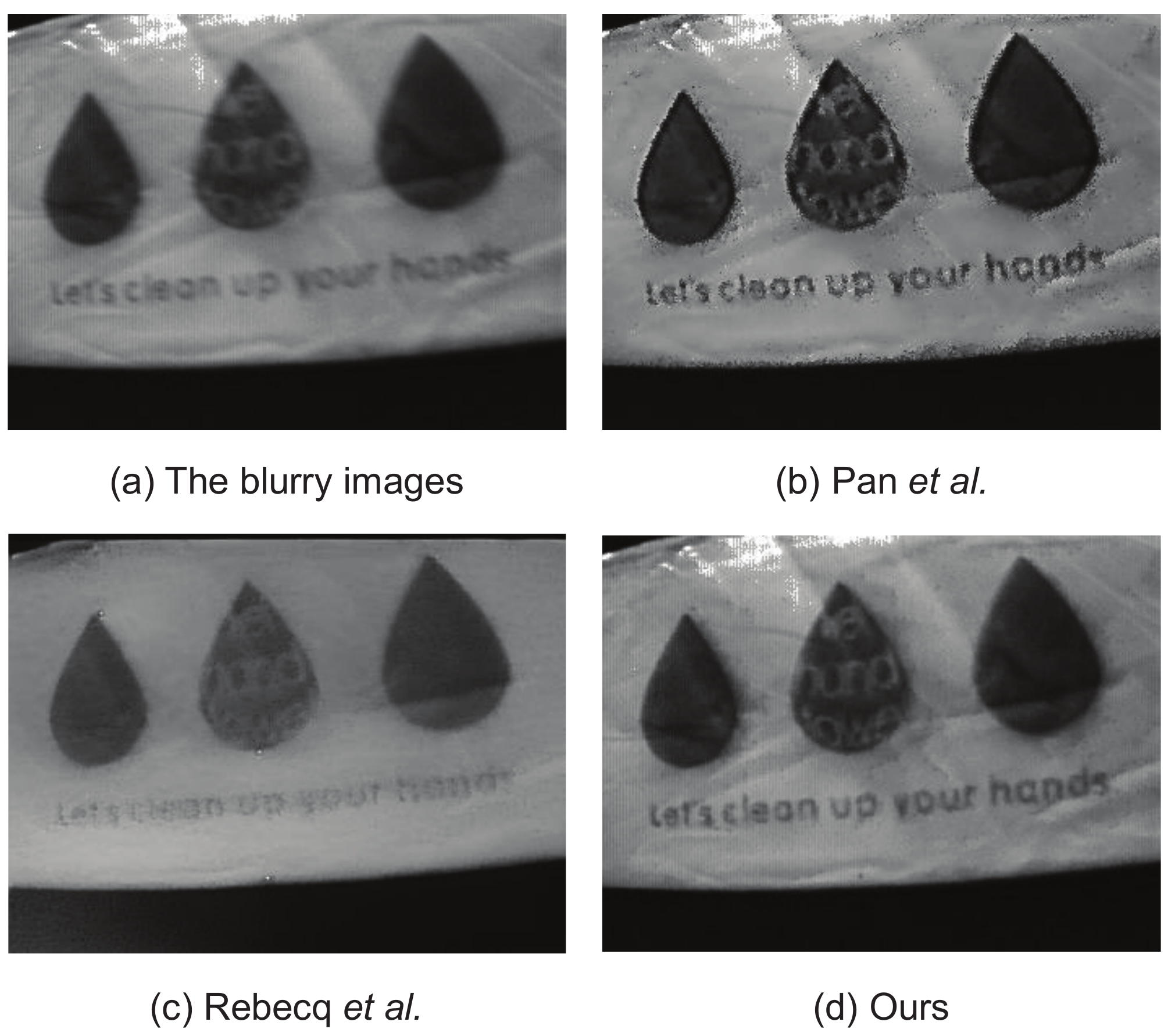}

   \caption{An example of restoring a sharp image from a blurry one (a) by Pan \etal~\cite{Pan2019EDI} (b), Rebecq \etal~\cite{Rebecq2019Highspeed} (c) and our method (d). Artifacts such as noise around the restored alphabets (b) and oversmooth of the folds (c) could be observed.
   }
\label{fig:discussion}
\end{figure}

\begin{table*}
\caption{Quantitative results using synthetic data containing both blurry images and event stacks. All the methods listed are tested under the same blurry condition. We have one deblurring baseline and two video generation baselines for comparison. Baseline-d and baseline-v1 do not use the residual learning method, and baseline-v2 does not use Conv-LSTM block. Although baseline-v2 has a slightly higher SSIM compared to our method, the visual quality of its generation shows a rather unstable appearance. Please refer to the supplement video for detailed comparison.}
\begin{center}
\begin{normalsize}
\begin{tabular}{c|c|c|c|c|c|c|c}
\hline
        \multicolumn{8}{c}{Average result of deblurring on synthetic dataset}\\
        \hline
&Nah \etal~\cite{Nah2017Deep}&Tao \etal~\cite{Tao2018Scale}&Kupyn \etal~\cite{Kupyn_2019_ICCV}&Purohit \etal~\cite{purohit2019bringing}&Pan \etal~\cite{Pan2019EDI}&Baseline-d &Ours \\
\hline
PSNR &31.30& 30.54&29.29& 30.58 &30.68&27.67&\textbf{32.99} \\
\hline
SSIM &0.9113& 0.924&0.8990& \textbf{0.9410} &0.9088&0.8995&0.9353\\
\hline
\end{tabular}
\\
\begin{tabular}{c|c|c|c|c|c|c|c}
\hline
    \multicolumn{8}{c}{Average result of HFR video generation on synthetic dataset}\\
    \hline
    &Jin \etal~\cite{Jin2018Learning}
    &Scheerlinck \etal~\cite{Scheerlinck2018Continuous}&Nah \etal~\cite{Nah_2019_CVPR}
    &Pan \etal~\cite{Pan2019EDI}&Baseline-v1&Baseline-v2&Ours\\
    \hline
    PSNR &28.01 &25.84&29.97 &28.06 &26.75 &31.52 &\textbf{31.76}\\
    \hline
    SSIM &0.8670 &0.7904&0.8947 &0.8623&0.8424&\textbf{0.9252} &0.9202\\
\hline
\end{tabular}
\end{normalsize}
\end{center}
\label{table:comparison table}
\end{table*}

\subsection{Relationship with existing models}\label{sec:comparison}
The EDI model described in~\cite{Pan2019EDI} can be treated as a special case of our model, by setting all $C_i$ as constant. It formulated the deblurring task with events to be a non-convex optimization problem and solved the constant threshold to restore the sharp image. While as we described above, the threshold in the physical model of the event camera is not a constant and the constant global threshold will lead to error and generate artifacts in the restored image, as \Fref{fig:discussion}(b) shows.

The learning-based model proposed in~\cite{Rebecq2019Highspeed} also used events to generate HFR video. While they used only events as input, lacking background and texture information in the generated video. Our residual net combines the restored clear background from intensity image with events and can generate a sharper video with richer background details, as \Fref{fig:discussion}(c) shows. More comparisons will be listed in the experiment section.

\section{Experiments and Evaluation}
In this section, we compare our method with state-of-the-art methods on image deblurring and HFR video generation. To qualitatively and quantitatively evaluate our method, we test our method on both synthetic and real datasets. 
\subsection{Data preparation}
\paragraph{Synthetic dataset.} In order to quantitatively compare our experiment result with other deblurring methods, we generate a synthetic dataset based on the GoPro dataset~\cite{Nah2017Deep}. This dataset consists of 240 FPS videos taken with GOPRO camera and then averaged varying number (7 $\sim$ 13) of successive latent frames to produce blurs of different strengths. To make a fair comparison, we also average 7 images to generate a blurry image as done in~\cite{Nah2017Deep}. And we generate events between two images using the simulator provided by Rebecq \etal~\cite{Rebecq18corl}.

\paragraph{Real dataset.} The dataset we use to evaluate our method is from~\cite{Pan2019EDI}, in which frames and events are captured by a DVS240 event camera, and consists of different scenes and motion patterns (\textit{e}.\textit{g}., camera shake, objects motion) that naturally introduce motion blur into the intensity images. We also record a dataset using a DVS346 event camera. Our dataset will be released with our code.

\subsection{Training details}\label{sec:training details}
Our network is trained on an NVIDIA Titan X GPU and is implemented on Tensorflow platform. The optimizer we choose is ADMM, and its learning rate is the constant 0.002. In our experiment, we find that 50 epochs are enough for our model to converge. During each epoch, we use a batch of 8 blurry images and related events. 

It is due to only blurry images existing in the real dataset that we only use the synthetic dataset to train our network. The GoPro dataset is divided into two parts with a proportion of $2:1$. During the training procedure, an image of 720$\times$1280 will be randomly cropped to 256$\times$256, having enough events in that area. Moreover, we convert pictures in the GoPro dataset to grayscale images, because that there are only grayscale images in the real dataset.

Moreover, we divide the input events into 6 bins, which means we make the input events 6 channels rather than just stack them into 1 channel because according to~\cite{Mohammad2019EventHDR}, more channels work better than simply 1 channel. In our experiment, 6 channels have shown higher restoration quality, which is 32.99dB vs.~29.93dB on PSNR and 0.9353 vs.~0.9043 on SSIM. 

\begin{figure*}
\centering
\includegraphics[width=1\textwidth]{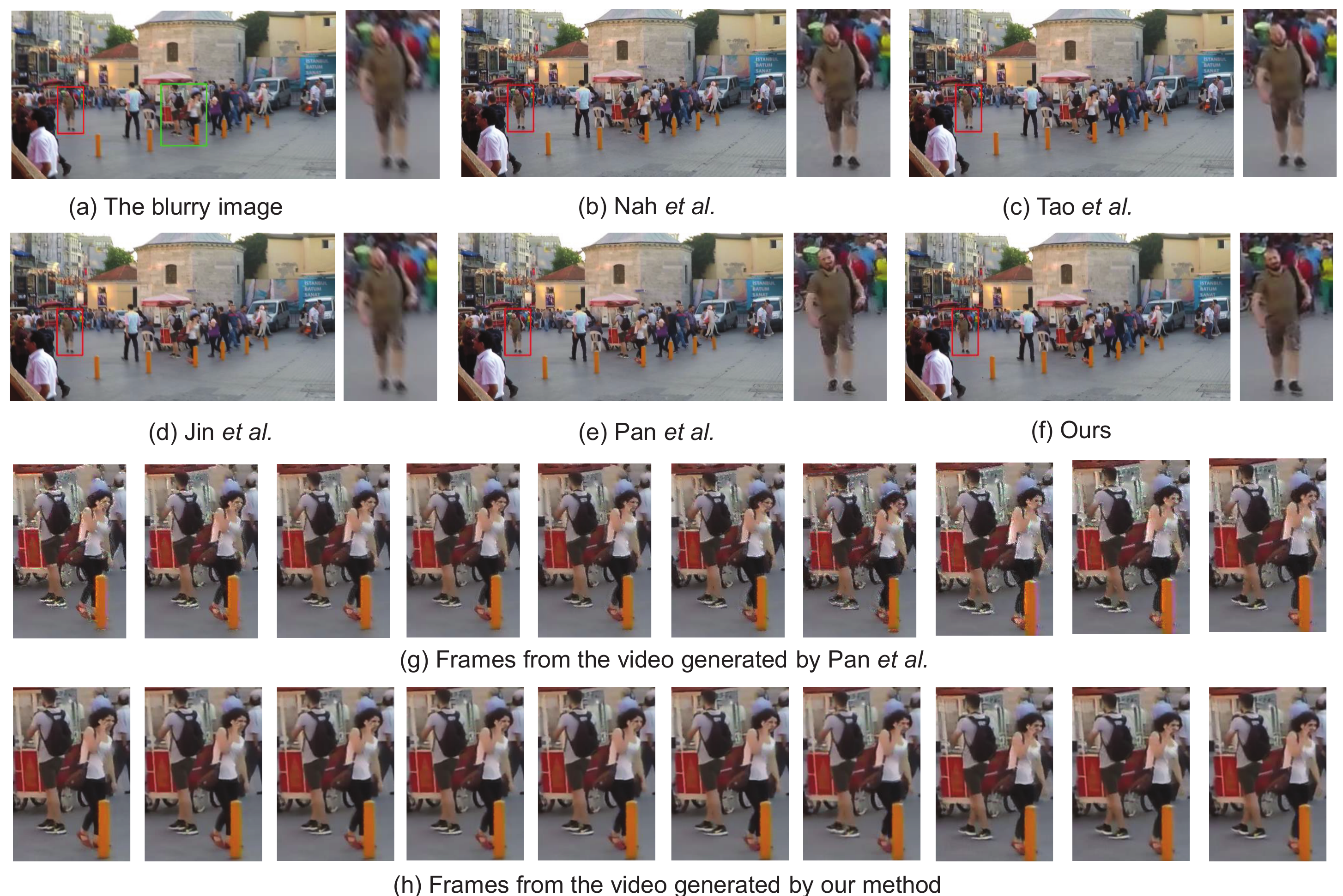}
   \caption{An example of deblurring and HFR video generation results on our synthetic event dataset based on the GoPro dataset~\cite{Nah2017Deep}. (a) The input blurry image. (b)$\sim$(f) Deblurring results of Nah \etal~\cite{Nah2017Deep}, Tao \etal~\cite{Tao2018Scale}, Jin \etal~\cite{Jin2018Learning}, Pan \etal~\cite{Pan2019EDI} and our method. (g) and (h) Frames from the generated video by Pan \etal~\cite{Pan2019EDI} and our method.}
   
\label{fig:comparison fig}
\end{figure*}

\subsection{Experimental results}\label{sec:exp result}
\paragraph{Deblurring.} We compare our proposed method with state-of-the-art learning-based image deblurring methods on synthetic dataset, including~\cite{Nah2017Deep,Kupyn_2019_ICCV,Tao2018Scale,ZhangPRSBL018}. We also compare with the result of EDI model~\cite{Pan2019EDI}. We generate synthetic events on each channel of the images, which is consistent with the principle of color DVS~\cite{Scheerlinck2019colorDVS}. Then we restore three channels separately and combine them for final output. To prove the effectiveness of the proposed residual network, we propose a baseline-d method without the global residual connection.

The deblurring results comparison is shown in \Tref{table:comparison table} and \Fref{fig:comparison fig}(a)$\sim$(f). As indicated in \Tref{table:comparison table}, our method achieves the best performance on PSNR and promising result on SSIM comparing to state-of-the-art methods. 

\begin{figure*}
\centering
\includegraphics[width=1\textwidth]{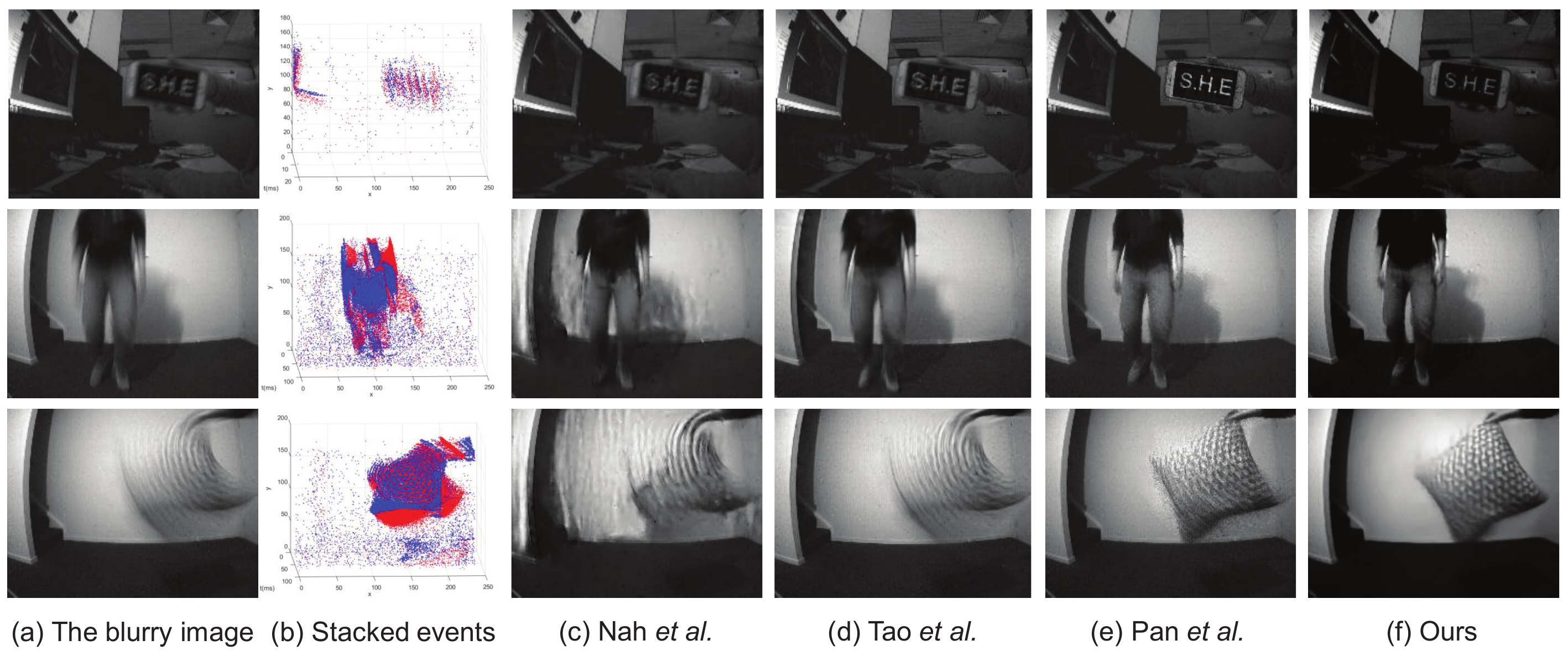}
\includegraphics[width=1\textwidth]{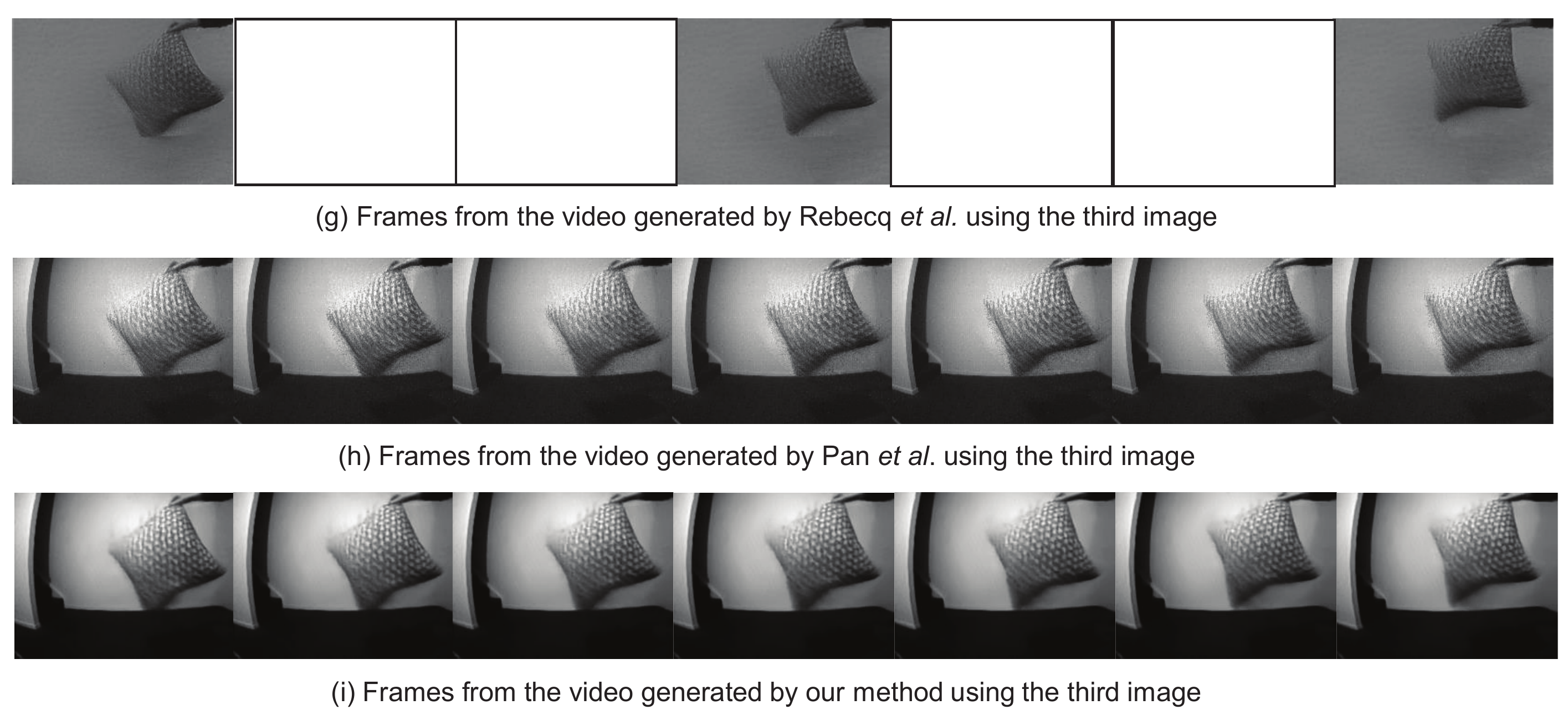}
   \caption{Examples of deblurring and HFR video generation results on the real dataset from~\cite{Pan2019EDI}. (a) The input blurry image. (b) The corresponding event data. (c)$\sim$(f) Deblurring results of Nah \etal~\cite{Nah2017Deep}, Tao \etal~\cite{Tao2018Scale}, Pan \etal~\cite{Pan2019EDI} and our method. (g)$\sim$(i) Frames of videos generated by Rebecq \etal~\cite{Rebecq2019Highspeed}\protect\footnotemark, Pan \etal~\cite{Pan2019EDI} and our method.
  }
\label{fig:real data example2}
\end{figure*}

We also compare our proposed method on a real dataset, whose results are shown in \Fref{fig:real data example1} and \Fref{fig:real data example2}. We can find clearly that our method restores a clearer image than the other learning-based methods. Comparing to the EDI model~\cite{Pan2019EDI}, our model suffers less from the background noise caused by the event camera, which leads to less noise in the restored image. 

\begin{figure}[h]
\centering
\includegraphics[width=0.45\textwidth]{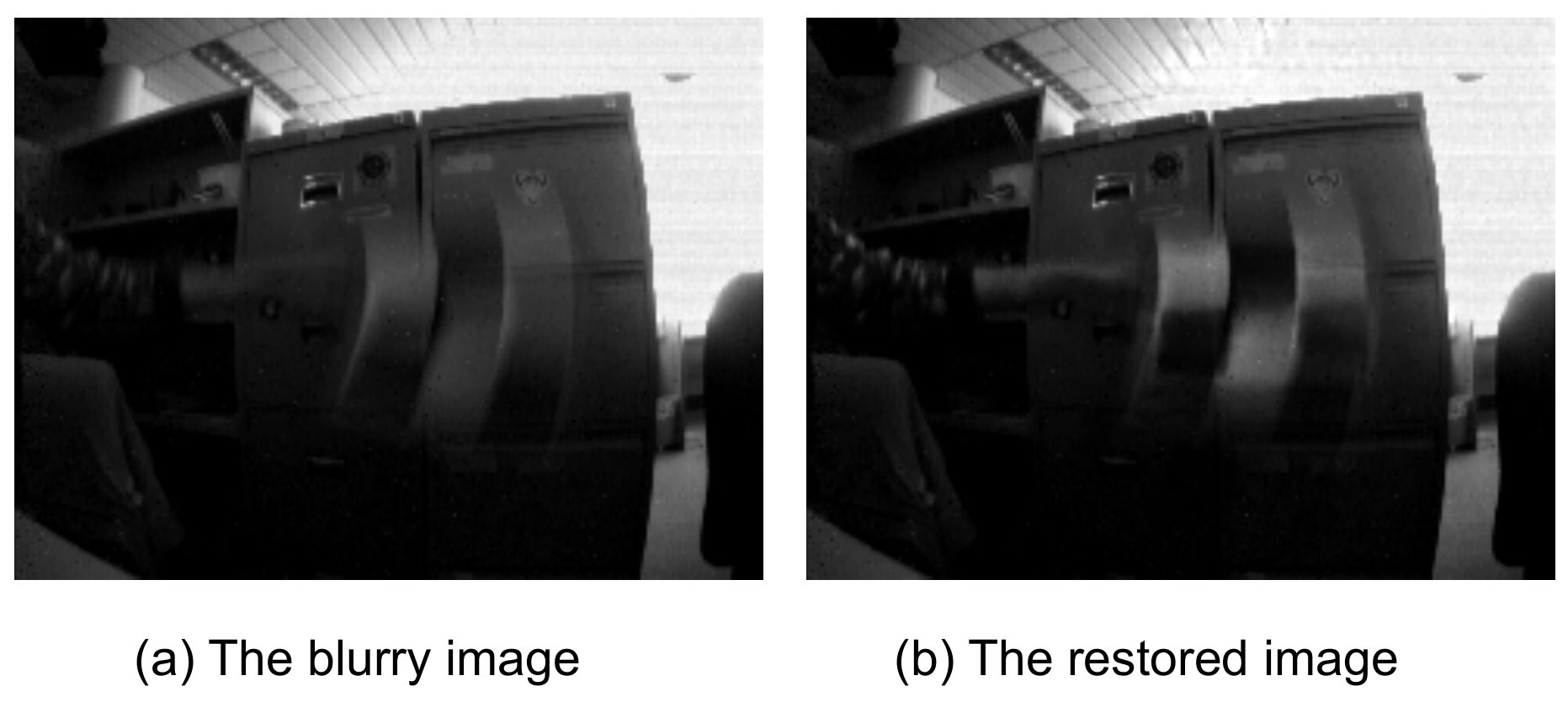}
   \caption{A failure example of our method, in which the checkerboard still remains blurry.}
\label{fig:failure_samples}
\end{figure}

\paragraph{HFR video generation.}We compare our video generation method with state-of-the-art generation methods~\cite{Scheerlinck2018Continuous,Jin2018Learning,Pan2019EDI,Rebecq2019Highspeed} in \Fref{fig:comparison fig}(g) and (h), \Tref{table:comparison table} for synthetic dataset and in \Fref{fig:real data example2} for real dataset. To prove the efficiency of our residual learning method to generate HFR video, we design two baseline methods: baseline-v1 without the global residual connection and baseline-v2 using Dense block in the downsampling procedure.

Compared with image-based method~\cite{Jin2018Learning}, our method achieves higher quality. In comparison to event-based generation methods, the video generated by our method has less noise than~\cite{Pan2019EDI} and richer details than~\cite{Rebecq2019Highspeed}. As for the baseline-v2, the quality of its output drops more rapidly than our method, though it has comparable results of PSNR and SSIM. Please refer to the supplement video for visual quality comparison.

\subsection{Failure samples}
There are indeed some pictures in the real dataset from ~\cite{Pan2019EDI} which our method cannot restore well, as shown in \Fref{fig:failure_samples}. It is due to that the event camera parameters in our synthetic dataset for training are different from those used in~\cite{Pan2019EDI}. The threshold and the bandwidth can both influence distribution of events. The steep intensity change like a checkerboard is not covered in our training data, resulting in poor performance as the example shown in \Fref{fig:failure_samples}.

\section{Conclusion}

We propose a residual learning deblurring method using event cameras, which has a modified U-Net structure with DenseNet blocks in each layer. Because of the high temporal resolution of event cameras, motion information is encoded in the output events. With events, our method can avoid restoring blurry images blindly as image-based methods and get more accurate results than existing event-based methods thanks to the effectiveness of the proposed residual representation and network. Using a similar structure with Conv-LSTM blocks, we can further generate an HFR video with a restored image and events.  

Our method does not get satisfying results when the event camera parameters are different from our synthetic data. The resolution of DVS is much lower (346$\times$260 and below) than an RGB camera. In our future work, we hope to combine high resolution color cameras with event cameras and construct a hybrid system to achieve higher resolution in either deblurring or HFR video generation tasks.

\footnotetext{For a better figure layout, we put placeholders where this method does not output a frame.}
{\small
\bibliographystyle{ieee}
\bibliography{final}
}

\end{document}